\pdfoutput=1

\documentclass[11pt]{article}

\usepackage[]{ACL2023}

\usepackage{times}
\usepackage{latexsym}

\usepackage[T1]{fontenc}

\usepackage[utf8]{inputenc}

\usepackage{microtype}

\usepackage{inconsolata}

\usepackage{algorithm}
\usepackage{algorithmic}
\usepackage[]{hyperref}
\usepackage{newfloat}
\usepackage{listings}

\usepackage{microtype}

\usepackage{amsmath}
\usepackage{amssymb}

\usepackage{float}
\usepackage{pgfplots}
\usepackage{subfigure}
\usepackage{booktabs}
\usepackage{multicol}
\usepackage{multirow}
\usepackage{arydshln} 
\usepackage{CJKutf8}
\usepackage{hyperref}

\title{Beyond Chain-of-Thought, Effective Graph-of-Thought Reasoning in Language Models}

\author{Yao Yao$^{1,2}$, Zuchao Li$^{3,*}$ and Hai Zhao$^{1,2,}$\thanks{$\ $  Corresponding author. This research was supported by the National Natural Science Foundation of China (No. 62306216), the Natural Science Foundation of Hubei Province of China (No. 2023AFB816), the Fundamental Research Funds for the Central Universities (No. 2042023kf0133), the Joint Research Project of
Yangtze River Delta Science and Technology Innovation Community (No.
2022CSJGG1400).} \\
$^{1}$Department of Computer Science and Engineering, Shanghai Jiao Tong University\\
$^{2}$MoE Key Lab of Artificial Intelligence, AI Institute, Shanghai Jiao Tong University\\
$^{3}$National Engineering Research Center for Multimedia Software, \\
School of Computer Science, Wuhan University, Wuhan, 430072, P. R. China \\
{\tt yaoyao27@sjtu.edu.cn, zcli-charlie@whu.edu.cn,}\\
{\tt zhaohai@cs.sjtu.edu.cn}\\
}

\begin{document}
\maketitle
\begin{abstract}

With the widespread use of language models (LMs) in NLP tasks, researchers have discovered the potential of Chain-of-thought (CoT) to assist LMs in accomplishing complex reasoning tasks by generating intermediate steps. However, human thought processes are often non-linear, rather than simply sequential chains of thoughts. Therefore, we propose Graph-of-Thought (GoT) reasoning, which models human thought processes not only as a chain but also as a graph. By representing thought units as nodes and connections between them as edges, our approach captures the non-sequential nature of human thinking and allows for a more realistic modeling of thought processes. GoT adopts a two-stage framework with an additional GoT encoder for thought graph representation and fuses the graph representation with the original input representation through a gated fusion mechanism. We evaluate GoT's performance on a text-only reasoning task (AQUA-RAT) and a multimodal reasoning task (ScienceQA). Our model achieves significant improvement over the strong CoT baseline on the AQUA-RAT test set and boosts accuracy from 85.19\% to 87.59\% using the T5-base model over the state-of-the-art Multimodal-CoT~\cite{DBLP:journals/corr/abs-2302-00923} on the ScienceQA test set. Our code is publicly available at \href{https://github.com/Zoeyyao27/Graph-of-Thought}{https://github.com/Zoeyyao27/Graph-of-Thought}

\end{abstract}

\section{Introduction}

\begin{figure*}[h]
    \centering
    \includegraphics[width=0.95\textwidth]{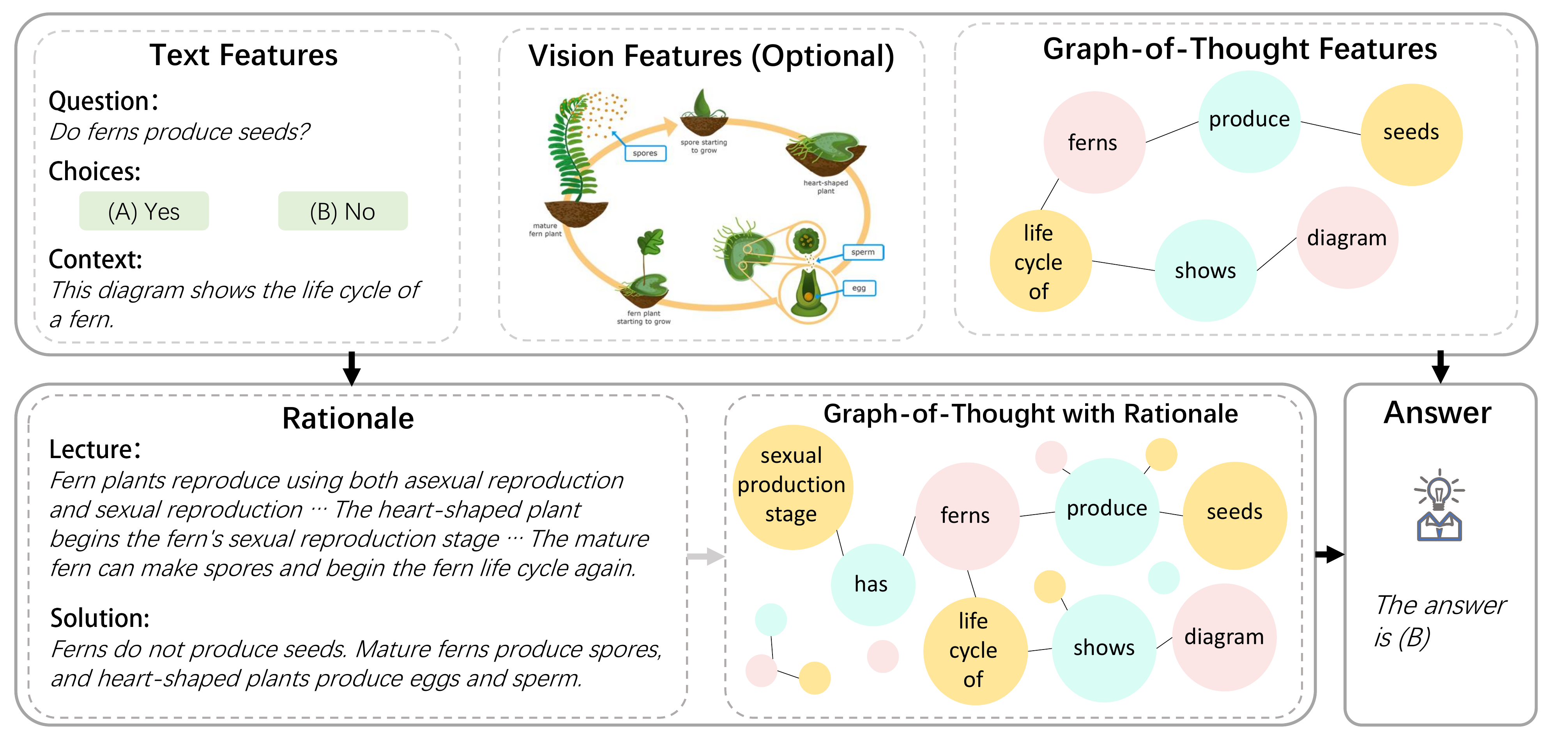}
    \caption{An example of GoT reasoning. Vision features are optional and are only required in multimodal reasoning.}
    \label{fig:example}
\end{figure*}

In the field of human cognition, it has long been recognized that the human thought process is far more complex and non-linear than could be captured by a simple, sequential chain of thoughts~\cite{barsalou1999perceptual}. Human thinking is often characterized by its ability to make sudden leaps and connections between seemingly unrelated ideas, which can lead to novel insights and solutions. This non-linear, jumping thought process is a hallmark of human creativity, reasoning, and problem-solving abilities. However, it also poses a significant challenge for cognitive modeling and understanding.

Recently, Large Language Models (LLMs) have been advancing at an unprecedented pace. With the emergence of breakthroughs such as GPT-3~\cite{DBLP:conf/nips/BrownMRSKDNSSAA20}, PaLM~\cite{DBLP:journals/corr/abs-2204-02311}, and GPT-4~\cite{GPT4}, the field of natural language processing has entered a new era of possibilities. Recent studies~\cite{DBLP:journals/corr/abs-2201-11903,DBLP:journals/corr/abs-2203-11171,DBLP:journals/corr/abs-2210-03493} have shown that the reasoning ability of LLMs can be unlocked by Chain-of-Thought (CoT) prompting. CoT prompting involves a series of intermediate natural language rationales that lead to the final answer. In addition,~\citet{DBLP:journals/corr/abs-2302-00923} have introduced Multimodal-CoT, which combines both language and visual modalities to help surpass the limitations of textual information. More detailed related works can be found in Appendix \ref{ap:related_work}.

Previous works on Chain-of-Thought (CoT) prompting, which have been limited to textual and visual information, often represented the human reasoning process as sequential thought chains. This approach overlooks the modeling of humans' jumping thought process and neglects to incorporate the complex structural information of reasoning thoughts into the model.
Concurrent work Tree-of-thoughts (ToT) \cite{yao2023tree} divides thoughts into thought units and models them as a tree-like search process. 

Nevertheless, human cognition transcends this tree structure, exhibiting intricate graph-like formations. Our perspective diverges further as we believe that the human intellect is capable of crafting elaborate thought graphs founded upon linear thoughts. Therefore, we aim to enable the concurrent assimilation of linear and nonlinear cognitive processes, surpassing the mere generation of segmented thought units. To address the above limitation, different from ToT, we propose the Graph-of-Thought (GoT), a novel approach to modeling human thought processes not only as a chain but also as a graph. Our method is based on the assumption that the human mind works by connecting and recombining ideas in a non-sequential, graph fashion, rather than following a strict sequential chain. By representing thought units as nodes and connections between thoughts as edges, GoT captures the rich, non-sequential nature of human thinking and allows for a more realistic and logical modeling of reasoning processes.

An example of GoT reasoning is shown in Figure \ref{fig:example}. Inspired by Multimodal-CoT~\cite{DBLP:journals/corr/abs-2302-00923}, we have adopted a two-stage reasoning framework. It first generates rationales and then generates the final answer based on the predicted rationales. In addition to text features, graph features of GoT are integrated during the rationale generation and answer inference. Specifically, GoT is first constructed with an Extract-Cluster-Coreference (ECC) process, which simulates the deductive process in human reasoning. We have used T5~\cite{2020t5}  pre-trained language model as our backbone model. GoT is encoded with a graph attention network and then fused with the original representation via a gated fusion network.

Furthermore, we have also presented a multimodal GoT, which integrates not only text features and GoT features but also visual features. For our experiments, we have used both FLAN-Alpaca \footnote{https://github.com/declare-lab/flan-alpaca. FLAN-Alpaca is developed by fine-tuning T5 model on the Flan collection} (T5)-base and FLAN-Alpaca (T5)-large as our backbone models. 

We implement GoT as a two-stage framework and fine-tuning language models and integrating text, thought graph, and vision features for a more realistic and accurate reasoning process. GoT demonstrates exceptional performance on both text-only AQUA-RAT ~\cite{ling-etal-2017-program} and multimodal ScienceQA~\cite{lu2022learn} benchmarks, surpassing the accuracy of online system ChatGPT~\cite{GPT4} by 9.28\%, strong baseline Multimodal-CoT~\cite{DBLP:journals/corr/abs-2302-00923} by 2.40\%, and even exceeding human performance, establishing a new state-of-the-art on ScienceQA test set with far fewer parameters.

\section{Graph-of-Thought}

\begin{figure*}[h]
    \centering
    \scalebox{0.95}{
    \includegraphics[width=1\textwidth]{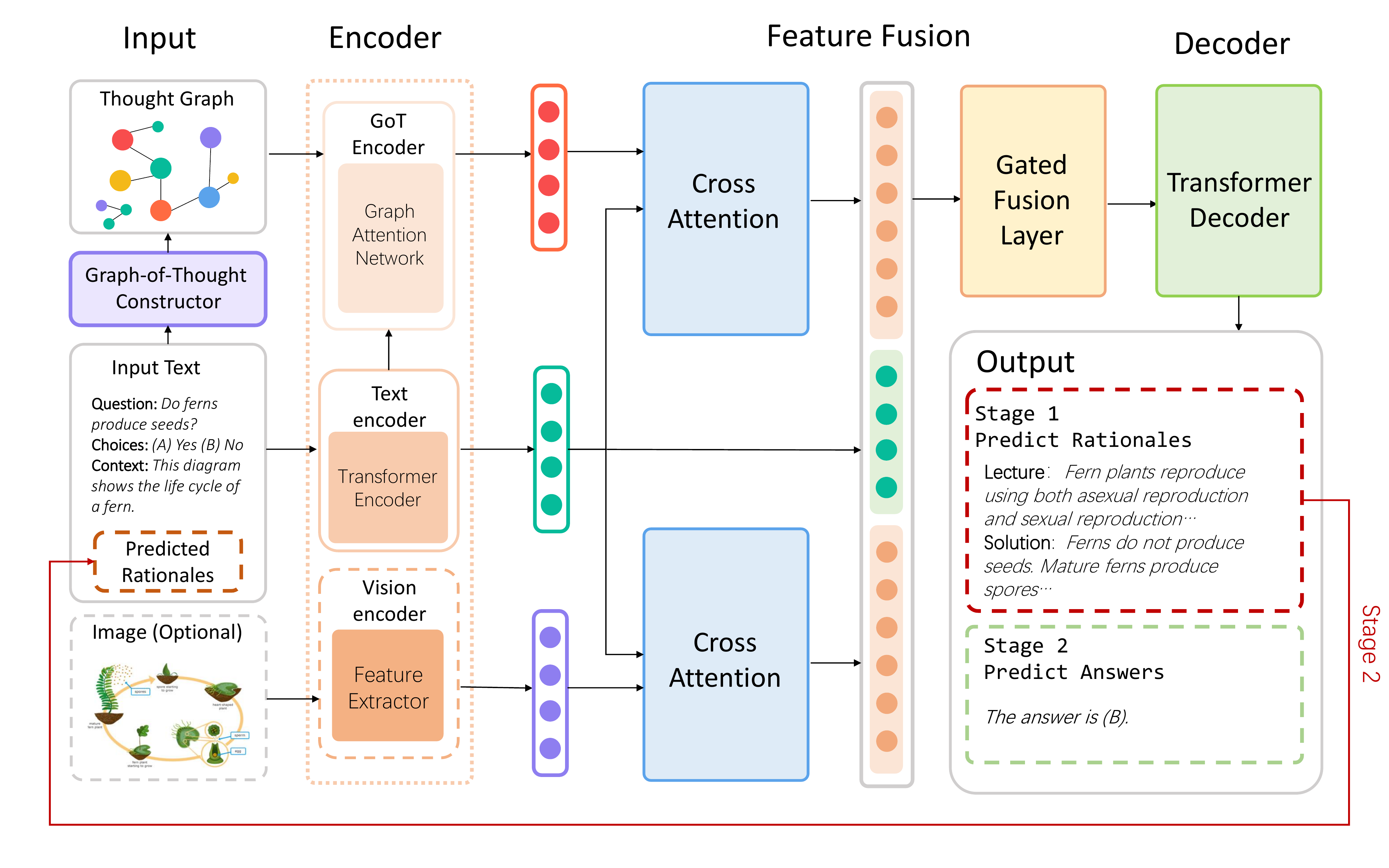}
    }
    \caption{Graph-of-Thought framework overview }
    \label{fig:model overview}
\end{figure*}

The overview of our proposed GoT can be seen in Figure \ref{fig:model overview}. Inspired by Multimodal-CoT~\cite{DBLP:journals/corr/abs-2302-00923}, GoT also adopts a two-stage framework.
(1) Rationale generation stage: In the first stage, the model generates rationales based on the input text (including question, context, and choices) the vision features, and the generated thought graph corresponding to the input text.
For multi-modal tasks \cite{DBLP:journals/corr/abs-2302-00923,DBLP:journals/corr/abs-2309-11436,DBLP:journals/corr/abs-2302-14045,DBLP:journals/corr/abs-2306-14824}, it is a common practice to use different encoders to process inputs from different modalities and a straightforward and versatile approach is to employ encoder-decoder models. Therefore, GoT employs independent encoders to encode input data for each modality. We use a Transformer encoder to encode input text, a vision encoder to encode an image, and a graph attention network to encode the thought graph. The encoded features are further passed into cross-attention to align text tokens with image patches and graph nodes, respectively. We then use a gated fusion layer to fuse these three features further and pass them into the Transformer decoder to predict the target rationales.
(2) Answer generation stage: The second stage aims at generating the final answer and is largely similar to the first stage. The main difference is that the input text is concatenated with the predicted rationales from the first stage.
It is worth noting that the above process describes a general multimodal reasoning framework. However, for text-only reasoning tasks, there are no image features, so the image encoding and vision feature fusion processes mentioned above can be omitted.
In the following section, we will provide a detailed exposition of the two key steps of our GoT reasoning framework: GoT construction and GoT encoding and feature fusion.

\subsection{GoT Construction}

\begin{figure*}[h]
    \centering
    \includegraphics[width=0.8\textwidth]{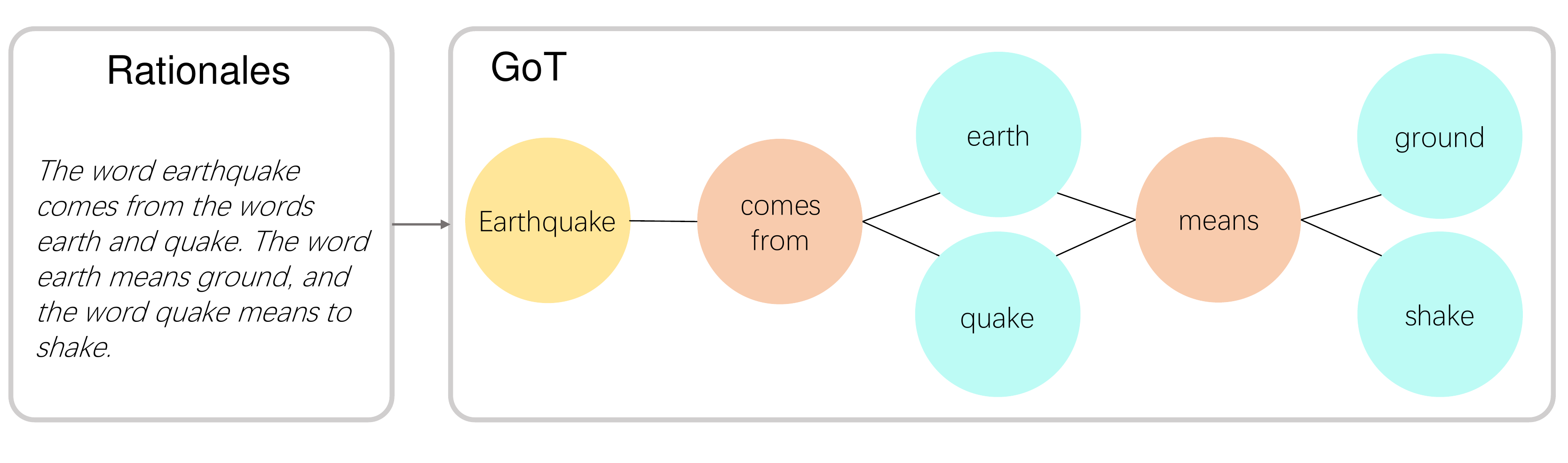}
    \caption{Graph-of-Thought deduction example }
    \label{fig:deduction_exapmle}
\end{figure*}

GoT employs thought graphs to simulate human deductive reasoning, thereby modeling humans' ability for leaps of thought. Our aim is to reflect the most fundamental deduction process by constructing a thought graph. If we have evidence that $x \rightarrow y$ and $y \rightarrow z$, then it follows that $x \rightarrow z$.
In Figure \ref{fig:deduction_exapmle}, the deduction reasoning can be formulated as follows: $\textit{Earthquake} \stackrel{\textit{comes from}}{\longrightarrow} \{\textit{earth, quake}\}$, $\{\textit{earth, quake}\} \stackrel{\textit{means}}{\longrightarrow} \{\textit{ground, shake}\}$. It is easy to reason that $\textit{Earthquake} {\longrightarrow} \{\textit{ground, shake}\}$.

We propose a novel Extract-Clustering-Coreference (ECC) process to construct thought graphs. ECC first extracts deductive triplets $T = \{t^i=(t^i_x,t^i_y,t^i_z)\}$ as the discrete raw graph, where $t^i_x$, $t^i_y$, and $t^i_z$ are thought units of the $i$-th triplet, and there exists an edge $e^i_{xy}$ between $t^i_x$ and $t^i_y$, and an edge $e^i_{yz}$ between $t^i_y$ and $t^i_z$.
Then, ECC clusters the nodes that refer to the same mentions to conduct coreference resolution. Specifically, we replace every graph node that belongs to a coreference cluster with the most representative mention in the cluster. By adopting this technique, our model is better equipped with denser thought graphs and the ability for deductive reasoning.
The detailed algorithm is illustrated in Algorithm \ref{alg1}.

In GoT construction, during the rationale generation stage, the input text consists of concatenated question, context, and choices. In multimodal GoT, image caption~\cite{lu2022learn} is appended to the input text for GoT to incorporate image information. During the answer inference stage, the predicted rationales from the rationale generation stage are further concatenated with the input text for corresponding GoT construction.

In our implementation of ECC process, inspired by~\cite{DBLP:conf/naacl/ChenY21}, we utilize open information extraction (OpenIE) systems \footnote{https://github.com/philipperemy/Stanford-OpenIE-Python}~\cite{angeli-etal-2015-leveraging} to extract subject-verb-object triplets as thought unit nodes. We apply coreference resolution to the extracted nodes using the Stanford CoreNLP system~\cite{manning-etal-2014-stanford}. The constructed thought graph is denoted as $\mathcal{G}(\mathcal{N},\mathcal{E})$, where $\mathcal{N}$ represents the nodes extracted by OpenIE and $\mathcal{E}$ represents the adjacency matrix. Rows and columns correspond to the nodes in the graph, and if there is an edge between two nodes, the corresponding matrix element is 1; otherwise, it is 0.

\renewcommand{\algorithmicrequire}{\textbf{Input:}}
\renewcommand{\algorithmicensure}{\textbf{Output:}}
\begin{algorithm}[H]
 \caption{ECC process}
\begin{algorithmic}
\REQUIRE  Input text $S$
\ENSURE Thought graph $\mathcal{G}(\mathcal{N},\mathcal{E})$
    \STATE Extract deductive triplet set $T$ from $S$ 
    \STATE $T=\{t^0,t^1,...,t^n\}$, $t^i=(t^i_x,t^i_y,t^i_z)$
    \FOR{ every triplet $t^i \in T$}
    \STATE $\mathcal{N}_{r} \gets \mathcal{N}_{r} \cup \{t^i_x,t^i_y,t^i_z\}$ 
    \STATE $\mathcal{E}_{r} \gets \mathcal{E}_{r} \cup \{e^i_{xy},e^i_{yz}\}$ 
    \ENDFOR
    \STATE extract coreference clusters $\mathcal{C}$ for $\mathcal{N}_{r}$
    \FOR{every node $n_i \in \mathcal{N}_r$}
    \IF {$n_i \in \forall c_j \in \mathcal{C}$}
    \STATE $n^*_j \gets \textrm{most representative mention in } c_j$
    \STATE $\mathcal{N} \gets \mathcal{N} \cup \{n^*_j\}$ 
    \ENDIF
    \ENDFOR
    \STATE Reconnect $\mathcal{N}$ based on $\mathcal{E}_r$ to construct $\mathcal{E}$
    \RETURN $\mathcal{N}$ , $\mathcal{E}$
\end{algorithmic} 
 \label{alg1}
\end{algorithm}

\subsection{GoT Encoding and Integration}

GoT reasoning utilizes separate encoders to encode input data for each modality. The thought graph is encoded using a graph attention network, while the input text is encoded using a Transformer encoder. In multimodal GoT reasoning, the image is encoded using an additional vision encoder.

\subsubsection{Base Encoder}


\paragraph{Text Encoder} For text representation, we use the Transformer encoder (e.g. T5~\cite{2020t5}) to encode the input text. Given input sentence $S = \{w_0, ..., w_l\}$, we extract the hidden states from the last layer of the Transformer encoder to obtain the text representation $H^T$:
\begin{equation}
\begin{aligned}
 H^T =\{h_0,h_1, ..., h_l\} =\mathbf{Encoder}_\textrm{text}(S)       
\end{aligned}
\end{equation}
where $h_i$ is the hidden representation of token $i$ and $l$ represents the length of the text input.

\paragraph{Vision Encoder (Optional)} 
For multimodal reasoning with vision modality, following~\cite{DBLP:journals/corr/abs-2302-00923}, we extract patch-level features of image $I$ using readily available vision extraction model as vision encoder $\mathbf{Encoder}_{vision}$ and then employ a trainable projection matrix $\mathbf{W_I}$ to project the extracted features into the vision representation $H^I$ which have the same shape with $H^T$.
\begin{equation}
 H^I = \mathbf{W_I}\mathbf{Encoder}_\textrm{vision}(I)
\end{equation}


\subsubsection{GoT Encoder}

\paragraph{Node Embedding} We first use special tokens {\tt <s>} and {\tt </s>} to highlight every thought graph node. Specifically, for node set with $j$ nodes $\mathcal{N}=\{n_0,...n_j\}$ , we construct the node input as $p$ and then feed the $p$ into the same text encoder and utilize the output representation of the special token {\tt <s>} as the initial node representation. Formally,
\begin{equation}
    p=[\textrm{\tt<s>}, n_0,\textrm{\tt</s>},...,\textrm{\tt<s>}, n_j, \textrm{\tt</s>}]
\end{equation}
\begin{equation}
    [h^s_0 ,h^n_0,h^e_0, ..., h^s_j, h^n_j,h^e_j]=\mathbf{Encoder}_\textrm{text}(p)
\end{equation}
where the $h^s_i$ and $h^e_i \in \mathbb{R}^{D}$ are the representation of {\tt <s>} and {\tt </s>} for node $n_i$ respectively, $D$ is the dimension of node embedding, and the $h^n_i = \{h^n_{i,1}, ..., h^n_{i,m}\}$ is the representations of node $n_i$ with $m$ tokens. we use the $h^s_i$ to represent the node representation of $n_i$.

\begin{figure}[H]
    \centering
    \includegraphics[width=0.4\textwidth]{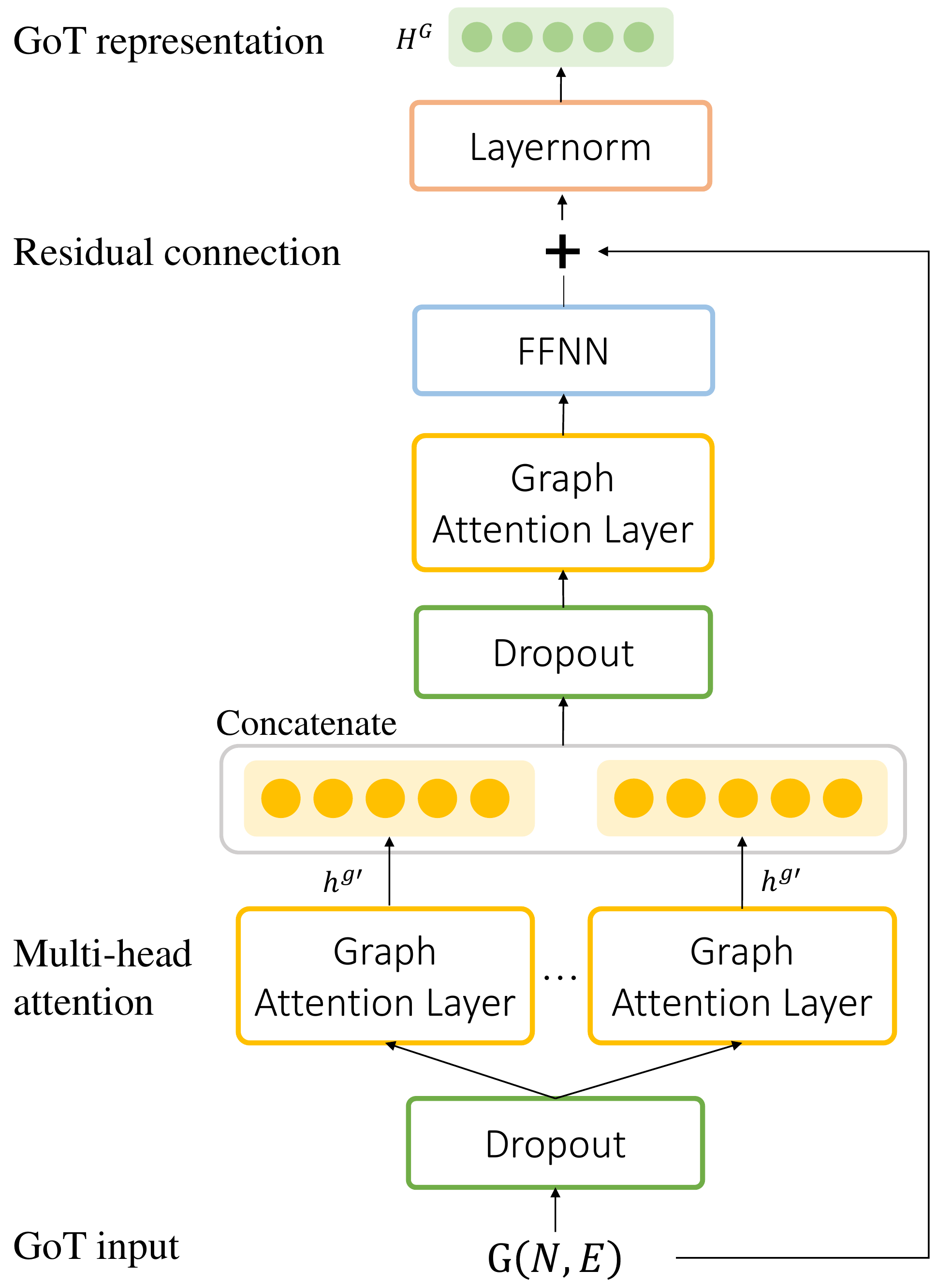}
    \caption{Architecture of GoT encoder}
    \label{fig:mmencoder}
\end{figure}

\paragraph{GAT Encoder} 
We employ a graph attention network (GAT)~\cite{DBLP:conf/iclr/VelickovicCCRLB18, DBLP:conf/naacl/ChenY21} to encode the thought graph. For every node $n_i$ in graph $\mathcal{G}(\mathcal{N},\mathcal{E})$, the \textbf{graph attention layer} is designed as:


\begin{equation} 
\begin{aligned}
a_{ij} = \mathbf{Attention} (\left[\mathbf{W} h^s_i||\mathbf{W} h^s_j\right] ) 
\end{aligned}
\end{equation}
\begin{equation} 
\begin{aligned}
q_{ij}=\mathbf{LeakyReLU}\left(a_{ij}\right)
\end{aligned}
\end{equation}


\begin{equation}
\begin{aligned}
\alpha_{i j}  =\mathbf{Softmax}(q_{ij}) =\frac{\exp \left(q_{ij}\right)}{\sum_{k \in \mathcal{K}_i} \exp \left( q_{ik} \right)}  
\end{aligned}
\end{equation}
\begin{equation}
\begin{aligned}
 h^{g\prime}_i=\mathbf{GELU}\left(\sum_{j \in \mathcal{K}_i} \alpha_{i j} \mathbf{W} h^s_j\right)
\end{aligned}
\end{equation}


where $||$ denotes concatenate operation, the $\mathbf{W}$ is a trainable weight and the set $\mathcal{K}_i$ contains the node $n_i$'s neighbours in thought graph $\mathcal{G}$. Our graph attention layer first employed a shared attention mechanism $\mathbf{Attention}(.): \mathbb{R}^{D^{\prime}} \times \mathbb{R}^{D^{\prime}} \rightarrow \mathbb{R}$ to compute the attention weights, where $D^{\prime}$ is the attention layer output dimension. The attention weights $a_{ij}$ measures the importance of node $n_i$'s features to $n_j$'s features. By only calculating the attention weights between nodes who are neighbours, our graph attention layer demonstrates the ability to perceive structural information of graphs. In our implementation, we adopt a single-layer feed-forward neural network (FFNN) as the attention mechanism which is both simple and straight-forward.

Figure \ref{fig:mmencoder} shows the architecture of our GoT encoder. Our GoT encoder employs a multi-head graph attention layer, following~\cite{DBLP:conf/iclr/VelickovicCCRLB18}, we concatenate the output of each graph attention layer and further pass it to a output graph attention layer with the same architecture:
\begin{equation}
   h_i^{g\prime}=\|_{k=1}^K \mathbf{GELU} \left(\sum_{j \in \mathcal{N}_i} \alpha_{i j}^k \mathbf{W}^k h^s_j\right)
\end{equation}
\begin{equation}
   h_i^{g\prime\prime}= \mathbf{GELU} \left(\sum_{j \in \mathcal{N}_i} \alpha_{i j} \mathbf{W} h^{g\prime}_j\right)
\end{equation}
where $K$ is the number of attention heads, $||$ is the concatenate operation, and $n$ is the number of nodes in thought graph. We then use a single-layer feed-forward neural network (FFNN) to obtain the final thought graph embedding $H^G$:

\begin{equation}
    h^{g\prime\prime}=[h_0^{g\prime\prime}, ..., h_n^{g\prime\prime}]
    ;\qquad
    H^G=\textbf{FFNN}(h^{g\prime\prime})
\end{equation}

\subsection{Feature Fusion}
After obtaining the encoded features, we use a single head attention to align the text representation $H^T$ with image representation $H^I$ and thought graph representation $H^G$, respectively. The image attention output $\mathbf{H^I}$ and thought graph attention output $\mathbf{H^G}$ are calculated by:

\begin{equation}
    \begin{aligned}
        \mathbf{H^I} =\mathbf{Softmax}\left(\frac{H^T H^{I\top}}{\sqrt{d}}\right)H^I 
    \end{aligned}
\end{equation}
\begin{equation}
    \begin{aligned}
        \mathbf{H^G}=\mathbf{Softmax}\left(\frac{H^T H^{G\top}}{\sqrt{d}}\right)H^G
    \end{aligned}
\end{equation}

where $Q$ is $H^T$ and $d$ is the dimension of $H^T$. We take both $K_I$ and $V_I$ as $H^I$ and $K_G$ and $V_G$ as $H^G$.
Please note that image representation is optional and is only required for multimodal dataset.

Next, a gated fusion mechanism~\cite{wu-etal-2021-good,DBLP:journals/corr/abs-2302-00923,li-etal-2022-vision,DBLP:conf/iclr/0001C0USLZ20} is applied to combine the attention outputs $\mathbf{H^I}$ and $\mathbf{H^G}$ with the text representation $H^T$. The feature fusion output $H$ can be calculated by:

$$
    \lambda=\left\{
    \begin{aligned}
    &\operatorname{Sigmoid}\left(W_T H^T+W_G\mathbf{H^G}\right) \\
    & \qquad \qquad \qquad \qquad \qquad \qquad \text{text-only}\\
    &\operatorname{Sigmoid}\left(W_T H^T+W_I\mathbf{H^I}+W_G\mathbf{H^G}\right) \\ 
    & \qquad \qquad \qquad \qquad \qquad \qquad \text{multimodal}
    \end{aligned}
    \right.
$$

$$
    H=\left\{
    \begin{aligned}
    &(1-\lambda) \cdot H^T+\lambda \cdot \mathbf{H^{G}} \quad \\
    & \qquad \qquad \qquad \qquad \qquad \qquad \text{text-only}\\
    &(1-\lambda) \cdot H^T+\lambda \cdot \mathbf{H^{I}}+\lambda \cdot \mathbf{H^{G}} \\
    & \qquad \qquad \qquad \qquad \qquad \qquad \text{multimodal}\\
    \end{aligned}
    \right.
$$


where $W_T$,$W_I$ and $W_G$ are all trainable weights. We then input the fused feature output $H$ into the decoder to predict the rationales or the final answer.

\section{Experiments}

\paragraph{Dataset}
We evaluate our model on the text-only AQUA-RAT~\cite{ling-etal-2017-program} and multimodal ScienceQA benchmark~\cite{lu2022learn}.  The detailed dataset information and statistics are shown in Appendix \ref{ap:data}.

\paragraph{Model Setup}

In our experiments, we used T5~\cite{2020t5} as our basic model architecture, including both T5-base and T5-large model sizes. Specifically, to ensure a fair comparison, we initialized our model with the finetuned T5 checkpoint FLAN-Alpaca \footnote{https://huggingface.co/declare-lab/flan-alpaca-base} and used ViT-large encoder~\cite{DBLP:conf/iclr/DosovitskiyB0WZ21} for the vision encoder, following~\cite{DBLP:journals/corr/abs-2302-00923}. We fine-tuned the models for 100 epochs with a learning rate of 5e-5. The detailed training parameters are available in Appendix \ref{ap:param}. We trained our models on four NVIDIA A800 80G GPUs.

\section{Results and Discussion}
\subsection{Main Results}

\paragraph{Baselines}

For AQUA-RAT, our baselines include: (1) Zero-Shot and Zero-Shot-CoT LLMs \cite{DBLP:journals/corr/abs-2205-11916}; (2) 
Few-Shot and Manual-CoT LLMs \cite{DBLP:conf/nips/Wei0SBIXCLZ22} and Auto-CoT \cite{DBLP:journals/corr/abs-2210-03493} (The above baselines all use the text-davinci-002 version of GPT-3 with 175B parameters); (3) Fintuned LLMs: Calcformer-T5-L \cite{kadlcik-etal-2023-calc} which finetunes calculator-using T5-Large model on the Calc-X collection. To have a fair comparison we also fine-tuned FLAN-Alpaca$_\textrm{base}$ and FLAN-Alpaca$_\textrm{large}$ on AQUA-RAT with traditional two-stage CoT.

For ScienceQA, following~\cite{DBLP:journals/corr/abs-2302-00923,lu2022learn}, our adopted baselines include: (1) Vision question answering (VQA) baseline models~\cite{DBLP:conf/cvpr/Yu0CT019,DBLP:conf/cvpr/00010BT0GZ18,DBLP:conf/nips/KimJZ18,DBLP:conf/cvpr/GaoJYLHWL19,DBLP:conf/icml/KimSK21,DBLP:conf/nips/LuQCXZZYLZ21,DBLP:journals/corr/abs-1908-03557,DBLP:conf/acl/LiYYHC20}; (2) Text-to-text LLMs~\cite{DBLP:journals/jmlr/RaffelSRLNMZLL20,DBLP:conf/nips/ChenKSNH20}  and (3) Text-to-text LLMs with CoT prompting~\cite{lu2022learn,DBLP:journals/corr/abs-2302-00923}. Both UnifiedQA~\cite{lu2022learn} and GPT-3.5~\cite{lu2022learn} use generated image captions to incorporate vision semantics. Whereas, Mutimodal-CoT~\cite{DBLP:journals/corr/abs-2302-00923} injects generated image features into traditional CoT reasoning. 

\paragraph{Results}
\begin{table}[h]
\centering

\scalebox{0.6}{
\begin{tabular}{lllc}
\toprule
MODELS           & TRAINING  & SIZE & ACC(\%) \\ \midrule
Zero-Shot \cite{DBLP:journals/corr/abs-2205-11916}   & zero-shot & 175B & 22.40 \\
Zero-Shot-CoT \cite{DBLP:journals/corr/abs-2205-11916}   & zero-shot & 175B & 33.50 \\
Few-Shot \cite{DBLP:conf/nips/Wei0SBIXCLZ22}   & few-shot & 175B & 24.80 \\
Manual-CoT \cite{DBLP:conf/nips/Wei0SBIXCLZ22}   & few-shot & 175B & 35.80 \\
Auto-CoT \cite{DBLP:journals/corr/abs-2210-03493}   & few-shot & 175B & 36.50 \\
 Calcformer-T5-L \cite{kadlcik-etal-2023-calc} & train-set & 770M & 27.20 \\
 \midrule
FLAN-Alpaca$_\textrm{base}$   & train-set & 223M & 30.09 $\pm$ 1.12    \\
\textbf{GoT-T5$_\textrm{base}$ }       & train-set & 223M & \textbf{32.09} $\pm$ 1.62    \\ \midrule
FLAN-Alpaca$_\textrm{large}$ & train-set & 738M & 33.73 $\pm$ 1.14     \\
\textbf{GoT-T5$_\textrm{large}$   }    & train-set & 738M & \textbf{34.48} $\pm$ 1.11     \\ \bottomrule
\end{tabular}
}
\caption{Main test accuracy results (ACC\%) of AQUA-RAT. Size=backbone model size. }
\label{tab:main-results-gsm}
\end{table}

\begin{table*}[h]
\centering
\scalebox{0.65}{
\begin{tabular}{lccccccccccc}
\toprule
MODEL               & TRAINING & SIZE & NAT            & SOC            & LAN            & TXT            & IMG            & NO             & G1-6           & G7-12          & AVG            \\ \midrule
Human               & -          & -    & 90.23          & 84.97          & 87.48          & 89.60          & 87.50          & 88.10          & 91.59          & 82.42          & 88.40          \\ \midrule
\multicolumn{12}{l}{\textit{Vision question answering baselines}}                                                                                                                                 \\
MCAN~\cite{DBLP:conf/cvpr/Yu0CT019}                & train-set          & 95M  & 56.08          & 46.23          & 58.09          & 59.43          & 51.17          & 55.40          & 51.65          & 59.72          & 54.54          \\
Top-Down~\cite{DBLP:conf/cvpr/00010BT0GZ18}            & train-set           & 70M  & 59.50          & 54.33          & 61.82          & 62.90          & 54.88          & 59.79          & 57.27          & 62.16          & 59.02          \\
BAN~\cite{DBLP:conf/nips/KimJZ18}                 & train-set           & 112M & 60.88          & 46.57          & 66.64          & 62.61          & 52.60          & 65.51          & 56.83          & 63.94          & 59.37          \\
DFAF~\cite{DBLP:conf/cvpr/GaoJYLHWL19}               & train-set           & 74M  & 64.03          & 48.82          & 63.55          & 65.88          & 54.49          & 64.11          & 57.12          & 67.17          & 60.72          \\
ViLT~\cite{DBLP:conf/icml/KimSK21}               & train-set           & 113M & 60.48          & 63.89          & 60.27          & 63.20          & 61.38          & 57.00          & 60.72          & 61.90          & 61.14          \\
Patch-TRM~\cite{DBLP:conf/nips/LuQCXZZYLZ21}          & train-set           & 90M  & 65.19          & 46.79          & 65.55          & 66.96          & 55.28          & 64.95          & 58.04          & 67.50          & 61.42          \\
VisualBERT~\cite{DBLP:journals/corr/abs-1908-03557,DBLP:conf/acl/LiYYHC20}          & train-set           & 111M & 59.33          & 69.18          & 61.18          & 62.71          & 62.17          & 58.54          & 62.96          & 59.92          & 61.87          \\ \midrule
\multicolumn{12}{l}{\textit{Text-to-text LLMs}}                                                                                                                                                  \\
UnifiedQA$_\textrm{base}$~\cite{DBLP:journals/jmlr/RaffelSRLNMZLL20}       & zero-shot          & 223M & 68.16          & 69.18          & 74.91          & 63.78          & 61.38          & 77.84          & 72.98          & 65.00          & 70.12          \\
GPT-3.5 ~\cite{DBLP:conf/nips/ChenKSNH20}           & zero-shot          & 175B & 74.64          & 69.74          & 76.00          & 74.44          & 67.28          & 77.42          & 76.80          & 68.89          & 73.97          \\   
\midrule
\multicolumn{12}{l}{\textit{Text-to-text LLMs with CoT}}                                                                                                                                         \\
UnifiedQA$_\textrm{base}$ (CoT)~\cite{lu2022learn} & zero-shot  & 223M & 71.00          & 76.04          & 78.91          & 66.42          & 66.53          & 81.81          & 77.06          & 68.82          & 74.11          \\
GPT-3.5 (CoT)~\cite{lu2022learn}      & 2-shot     & 175B & 75.44          & 70.87          & 78.09          & 74.68          & 67.43          & 79.93          & 78.23          & 69.68          & 75.17          \\

ChatGPT (CoT)~\cite{DBLP:journals/corr/abs-2304-09842}      &  few-shot    & - & 	78.82	&70.98	&83.18	&77.37	&67.92	&86.13	&80.72	&74.03  &78.31 \\
GPT-4 (CoT)~\cite{DBLP:journals/corr/abs-2304-09842}      &  few-shot    & - &  85.48	&72.44	&90.27	&82.65	&71.49	&92.89	&86.66	&79.04  &83.99\\\midrule

Mutimodal-CoT$_\textrm{base}$ ~\cite{DBLP:journals/corr/abs-2302-00923}   & train-set  & 223M & 84.37          & 88.30          & 84.36          & 83.72          & 80.32          & 86.90          & 85.83          & 84.05          & 85.19          \\
\multirow{2}{*}{\textbf{GoT-T5$_\textrm{base}$}}  & \multirow{2}{*}{train-set}  & \multirow{2}{*}{223M} & 86.25           & 93.55           & 85.51        & 85.89          & 86.30          & 86.34         & 87.79         & 87.23          & \textbf{87.59 }        \\
  &   &  & $\pm$ 0.31          & $\pm$ 0.06          & $\pm$ 0.11         & $\pm$ 0.32          &  $\pm$ 0.28          &  $\pm$ 0.12          &  $\pm$ 0.10          &  $\pm$ 0.40          &  $\pm$ 0.20          \\ \midrule
  
Mutimodal-CoT$_\textrm{large}$~\cite{DBLP:journals/corr/abs-2302-00923} & train-set  & 738M & 91.03          & 93.70          & 86.64          & 90.13          & 88.25          & 89.48          & 91.12          & 89.26          & 90.45          \\ 

\multirow{2}{*}{\textbf{GoT-T5$_\textrm{large}$}} & \multirow{2}{*}{train-set}  & \multirow{2}{*}{738M} & 90.88 & 93.57 & 88.45 & 90.26 & 88.16 & 90.29 & 91.19 & 90.14 & \textbf{90.81} \\ 
  &   &  & $\pm$ 0.22          & $\pm$ 0.38          & $\pm$ 0.44         & $\pm$ 0.35          &  $\pm$ 0.25          &  $\pm$ 0.47          &  $\pm$ 0.16          &  $\pm$ 0.23          &  $\pm$ 0.12          \\
  \bottomrule
\end{tabular}

}
\caption{Main test accuracy results (\%) of ScienceQA. SIZE=backbone model size. Question classes: NAT = natural science, SOC = social science, LAN = language science, TXT = text context, IMG = image context, NO = no context, G1-6 = grades 1-6, G7-12 = grades 7-12, AVG= average accuracy scores}
\label{tab:main-results}
\end{table*}

The rationales generation results can be seen in Table \ref{tab:rationale results} in Appendix \ref{append:rationale}. The overall results are reported in Table \ref{tab:main-results-gsm} and Table \ref{tab:main-results}.

In the AQUA-RAT dataset, our GoT$_\textrm{base}$ model attains a 0.78 enhancement in ROUGE-L scores for rationale generation during the initial stage, outperforming the FLAN-Alpaca$_\textrm{base}$ model, which does not integrate GoT. For the answer generation phase, the GoT$_\textrm{base}$ exhibits a substantial accuracy increase of 2.00\%, while the GoT$_\textrm{large}$ model records a 0.75\% enhancement. Compared to the 175B parameter zero-shot and few-shot LLMs, our GoT-large, employing just a 738M backbone model, achieves results remarkably close to those of Manual-CoT \cite{DBLP:conf/nips/Wei0SBIXCLZ22}.

For ScienceQA dataset, in rationale generation stage, we can see from Table \ref{tab:rationale results} that our model achieves a ROUGE-L of 94.39 and outperforms the Mutimodal-CoT$_\textrm{base}$ by 1.15.
For the final answer generation stage, our GoT achieves SOTA in all subjects and all grades. The most direct comparison is that our model achieves an accuracy of 87.59\% which is 2.40\% higher than that of the Mutimodal-CoT$_\textrm{base}$ with the similar number of parameters. 

GoT demonstrates a significant advantage over traditional CoT, elevating the accuracy from 30.09\% to 32.09\% in AQUA-RAT and from 85.19\% to 87.59\% in ScienceQA task. The results sufficiently suggest that utilizing thought graph features for deductive reasoning is a more effective approach than the existing methods, which only consider text or vision features by simply incorporating image captions or fusing generated image features.  In conclusion, our results confirm the effectiveness of utilizing two-dimensional graph-of-thought and demonstrate the potential of incorporating GoT into reasoning for LMs.

\subsection{Further Exploration}

\subsubsection{Ablation Study}

\paragraph{AQUA-RAT} In order to make sure that introducing thought graphs into GoT reasoning indeed boost the performance, we conduct the following experiments:

(1) \textbf{Random Thought Graph}

In the Random Thought Graph experiment, we maintain the GoT framework while introducing randomness into the process. We construct a thought graph by randomly selecting nodes and arbitrarily establishing connections between them. This approach is designed to evaluate the extent to which the GoT reasoning mechanism is reliant on the structured organization of thought graphs.
(2) \textbf{Triplets Concatenation}
In the Triplets Concatenation experiment, we take a straightforward approach by appending the extracted triplets directly to the input text. This method aims to assess the impact of omitting the structural information typically provided by thought graphs, offering insight into the significance of this structural element in the reasoning process.
(3) \textbf{Coreference Injection}

In the Coreference Injection experiment, we explore the potential benefits of integrating coreference resolution directly into the language model's reasoning process. We achieve this by incorporating coreference information into the input text, where all instances of coreferent entities are replaced with a consistent phrase, followed by model fine-tuning. This experiment seeks to understand the role of coreference resolution in enhancing the model's deductive capabilities.

\begin{table}[h]
\centering
\scalebox{0.75}{
\begin{tabular}{lcllll}
\toprule
\multicolumn{1}{c}{MODEL}   & \multicolumn{1}{l}{MODEL SIZE}  & \multicolumn{1}{c}{ACC} & \multicolumn{1}{c}{$\Delta$} \\ \midrule
\textbf{GoT-T5$_\textrm{base}$}                  & \multirow{2}{*}{233M}                     & \textbf{32.09}          & \multicolumn{1}{c}{-}     \\
\quad w/ Random Thought Graph        &                                                     & 30.31                  & -1.78                    \\ \midrule
Triplets Concatenation        &                233M                                     & 31.20                   & -0.89                  \\ \midrule
Coreference Injection        &                    233M                                 & 30.32                   & -1.77                  \\ 
\bottomrule
\end{tabular}
}
\caption{Ablation results of GoT on AQUA-RAT dataset.}
\label{tab:ablation}
\end{table}

Table \ref{tab:ablation} shows the overall ablation results. From the table, we can see that by randomly construct thought graphs to disrupt the deductive reasoning process, our model suffers a loss of 1.78\%, indicating the effectiveness of GoT. The results of Triplets Concatenation on the AQUA-RAT showed an accuracy of 31.20\%. This performance gap of 0.89 clearly demonstrates the significance of the structural information in our approach. For Coreference Injection, the model suffers a loss of 1.77 \% accuracy. We believe that these outcomes can be attributed to a couple of factors: (1) Simply replacing coreferent entities may lead to a loss of coherence in sentences, resulting in a reduction of semantic information and consequently having a limited impact on overall accuracy. (2) Open Information Extraction (OpenIE) for coreference resolution is not flawless, and direct replacement of entities might introduce noise that misleads the language model during judgment.

Contrastingly, the construction of a thought graph in the GoT framework does not compromise the original textual information (questions and rationales). Instead, it introduces additional structural assistance for LMs to conduct reasoning effectively. Thus, we contend that GoT's approach is indispensable and beneficial, as it supplements the LM's comprehension without introducing potential noise or loss of coherence in the input text.

\paragraph{ScienceQA}
To examine the impact of different backbone and vision encoder configurations on the GoT, we employed a distinct set of model settings. More specifically, we adopted the pre-trained T5 checkpoint UnifiedQA \cite{DBLP:conf/emnlp/KhashabiMKSTCH20} as the backbone model and utilized DETR \cite{DBLP:conf/eccv/CarionMSUKZ20} for the vision encoder, with results illustrated in the Table \ref{tab:different_model}. As shown, our GoT outperforms Mutimodal-CoT across various model configurations. A comparison reveals that GoT can achieve greater improvements on smaller models. We believe the main reason is that when the language model is not as robust, or when employing a relatively weaker vision encoder like DETR compared to ViT, GoT can leverage the inherent information within the language to enhance performance significantly. Additionally, to prove that our GoT's performance gain is not simply due to an increase in parameters, we conducted an ablation study. We expanded the parameter count of Multimodal-CoT$\textrm{base}$ to match our 233M model size by adding two layers of MLP instead of one in the gated fusion module, referred to as Multimodal-CoT$\textrm{base}$(enlarged). We also constructed a random thought graph ablation study on the ScienceQA dataset. The results from the ablation studies can be observed in the table \ref{tab:different_model}. From the table, it is evident that our model significantly outperforms the enlarged Multimodal-CoT by an accuracy of 2.04\%. These findings convincingly demonstrate the significance of incorporating thought graphs into multimodal reasoning. The performance of GoT with a randomly constructed thought graph was even lower than Mutimodal-CoT, indicating that when the language model and vision encoder are weaker, the model relies more heavily on GoT for reasoning.
\begin{table}[h]
\centering
\scalebox{0.8}{
\begin{tabular}{lrc}
\cline{1-3}
Model                                           & \multicolumn{1}{l}{ACC} & $\Delta$                         \\ \cline{1-3}
\multicolumn{3}{l}{\textit{UnifiedQA+DETR}}                                                             \\
Mutimodal-CoT$_\textrm{base}$                   & 77.67                   & -                            \\
Mutimodal-CoT$_\textrm{large}$                  & 81.37                   & -                             \\
\textbf{GoT$_\textrm{base}$}                    & \textbf{81.21}          & \multicolumn{1}{r}{3.54}   \\
\textbf{GoT$_\textrm{large}$}                   & \textbf{82.74}          & \multicolumn{1}{r}{1.37}    \\ \cline{1-3}
\multicolumn{3}{l}{\textit{Ablation Studies}}                                                                  \\
Mutimodal-CoT$_\textrm{base}$(enlarged) & 79.17                   & \multicolumn{1}{r}{-2.04}  \\
GoT$_\textrm{base}$ w/ Random Thought Graph      & 76.74                   & \multicolumn{1}{r}{-4.47}   \\ \cline{1-3}
                                                & \multicolumn{1}{l}{}    &                          
\end{tabular}
}
\caption{Ablation results of GoT on ScienceQA dataset. For GoT models $\Delta$ indicates the performance gains of GoT models over their Multimodal-CoT counterparts. In the ablation studies, $\Delta$ represents improvements relative to the GoT$_\textrm{base}$ model }
\label{tab:different_model}
\end{table}

\subsubsection{Analysis}

\paragraph{Performance on Different Classes}

\begin{figure}[htbp]
\centering
\scalebox{1}{
    \includegraphics[width=7cm]{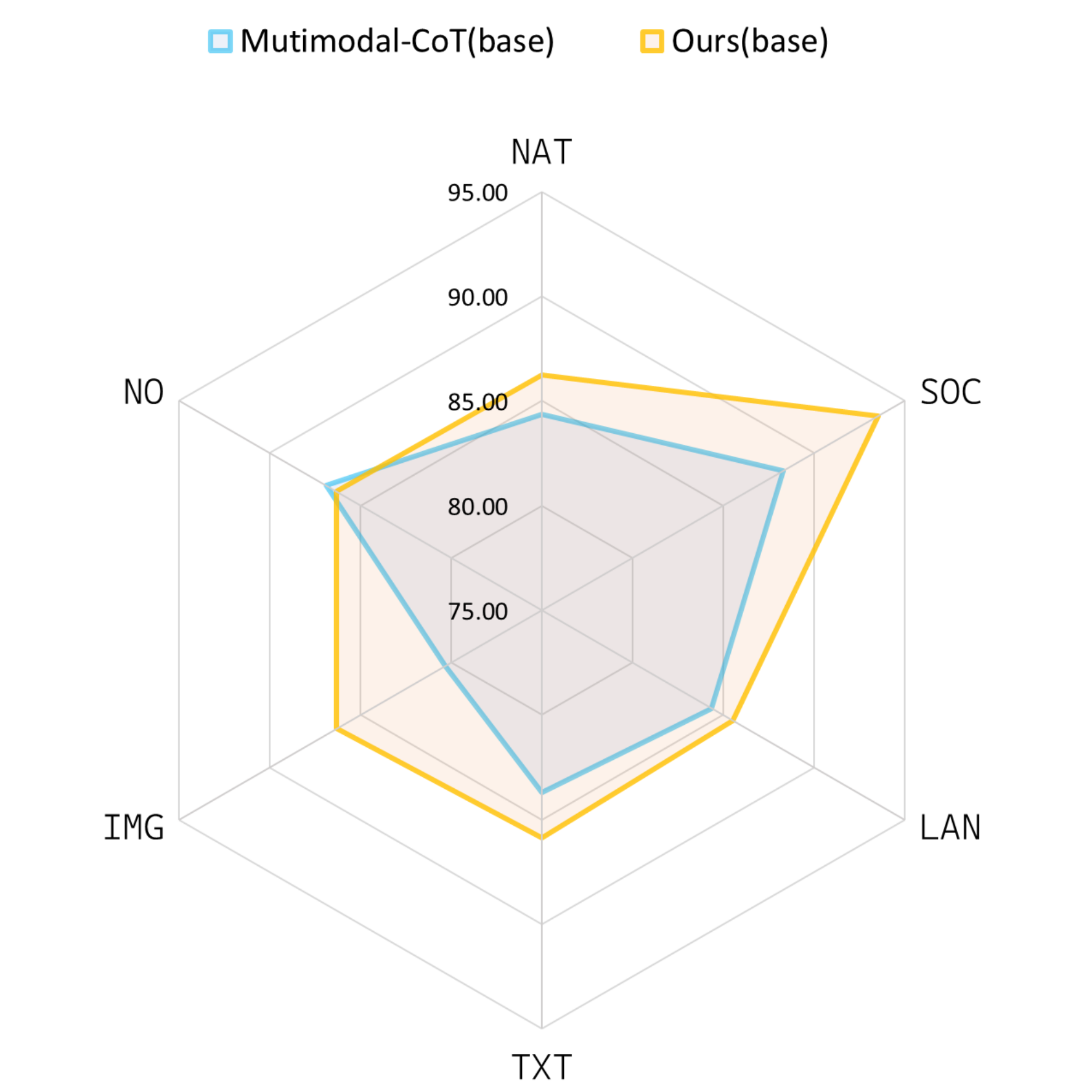}
}
    \caption{Performance on different question classes}
    \label{fig:radar}
\end{figure}

In order to investigate the impact of GoT on the overall model performance across different subjects , we calculated the accuracy for different subjects and compared it with that of Mutimodal-CoT. We also compare the performance of two models on different question classes.The radar Figure \ref{fig:radar} shows the overall results for our base model. With respect to various subjects and question classes, our model demonstrates superior performance over the Mutimodal-CoT$_\textrm{base}$ and attains a more consistent and enhanced outcome. Our model presents outstanding advantages especially in the field of social science, with an accuracy improvement of 5.25\%. For different question classes, our model demonstrates the largest improvement on questions involving images. Our hypothesis is that by constructing a thought graph and integrating the three features of text, image, and thought graph, we can better align the textual and visual information for the model, thus maximizing the utilization of visual information and obtaining more accurate answers.

\begin{figure}[htbp]
\centering
\scalebox{0.8}{
    \begin{tikzpicture}
        \begin{axis}[            xlabel={Grades},            ylabel={Accuracy(\%)},            legend style={at={(0.5,-0.35)},anchor=north,font=\small},  width=8cm,        height=7cm,          xmin=0.5, xmax=12.5,            ymin=60, ymax=100,            minor tick num=2,            grid=both,   grid style=dashed,         ]
            \addplot[orange,mark=*,line width=1pt] coordinates {
                (1,70.59)
                (2,82.59)
                (3,85.11)
                (4,89.75)
                (5,89.13)
                (6,91.29)
                (7,87.99)
                (8,85.55)
                (9,95.90)
                (10,93.28)
                (11,80.00)
                (12,88.37)
            };
            \addplot[cyan,mark=square,line width=1pt] coordinates {
                (1,76.47)
                (2,80.55)
                (3,85.42)
                (4,88.71)
                (5,85.71)
                (6,87.15)
                (7,85.34)
                (8,84.59)
                (9,94.06)
                (10,86.55)
                (11,80.49)
                (12,87.21)
            };

            \legend{Ours$_\textrm{base}$,Mutimodal-CoT$_\textrm{base}$}
        \end{axis}
    \end{tikzpicture}
    }
    \caption{Performance on different grades}
    \label{fig:line_chart}
\end{figure}

\paragraph{Performance on Different Grades}
It can be seen from the Table \ref{tab:main-results} that Mutimodal-CoT experience a decrease in accuracy of 1.78 as the grade level of the given question increases while GoT only has minor decrease of 0.56. We believe the main reason is that by incorporating GoT, models acquires the ability for deductive reasoning and can better comprehend the relationships between different entities and thus better understand the meaning of the problems. Through this method, for higher-grade problems with greater complexity, the model can construct a thought graph to help itself generate a more complete logical chain for deduction, thereby generating more accurate answers. More detailed model performance on different grades can be found in Figure \ref{fig:line_chart}. We can see that in the lower grade, two models achieves a similar performance. As the grade level increases and the difficulty of the questions becomes more challenging, the gap between our model and the Mutimodal-CoT model gradually widens. Due to the small number of questions ($\le 130$) available for each grade in grade 1 and grades 11-12, there is greater fluctuation in the accuracy of both models. Nevertheless, it is evident from the table that our model exhibits stronger and more stable advantages over Mutimodal-CoT in each grade. 

\paragraph{Case Study and Limitation} In order to gain a deeper understanding of the performance of GoT, we conduct case studies which can be found in the Appendix \ref{ap:case}. We also visualize the attention weights $a_{ij}$ in GoT encoder to demonstrate how GoT performs deductive reasoning to generate more accurate answers in Appendix \ref{ap:viz}. For the limitation of this work, compared to CoT, GoT may result in additional computational costs and slightly slower training times. Detailed limitation analysis can be found in Appendix \ref{ap:limiation}.

\section{Conclusion}

We introduce a novel Graph-of-Thought (GoT) reasoning approach, which is an innovative method for modeling the non-sequential nature of human thinking for LMs. GoT enhances LMs with deductive reasoning abilities, providing a more realistic representation of thought processes. Our experiments showcases the superiority of GoT on the text-only reasoning dataset AQUA-RAT, achieving a similar result compared to GPT-3 model while utilizing significantly fewer parameters. Furthermore, GoT establishes a new state-of-the-art on the multimodal reasoning benchmark, ScienceQA with fewer parameters. This performance surpasses strong ChatGPT and GPT-4 systems, as well as human performance, demonstrating the efficacy of GoT. Through comprehensive case studies and ablation studies, we provide substantial evidence of the effectiveness of GoT in reasoning tasks. If you want it, you GoT it!

\bibliography{custom}
\bibliographystyle{acl_natbib}

\clearpage
\appendix
\section*{Appendix}

\section{Related Works}
\label{ap:related_work}

In chain-of-thought reasoning, one idea leads to the next in a logical sequence and builds on previous knowledge. Each idea is supported by evidence or reasoning, and the conclusions drawn from the chain are logical and sound. Most CoT methods can be divided into two categories based on how to generate the final answer: (1) prompting for CoT, including zero-shot CoT and few-shot CoT; and (2) fine-tuning for CoT.

\paragraph{Zero-shot CoT Prompting}

As large language models continue to advance rapidly, many researchers are beginning to explore CoT reasoning for LLMs. The zero-shot CoT method proposed by~\citet{DBLP:journals/corr/abs-2205-11916} consists of two stages: (1) adding a "\textit{Let's think step by step}" prompt to generate CoT, and (2) concatenating the generated CoT and adding the phrase "\textit{So the answer is}" to obtain the final answer.
Tree-of-Thought (ToT) \cite{yao2023tree} enables deliberate decision-making through exploration of coherent text units. ToT divides thoughts into thought units and models them as a tree-like search process. Although both GoT and ToT aim to capture human non-linear thoughts, GoT is distinct from ToT in terms of both methodology and objectives. We believe that human thinking involves both linear and non-linear aspects. Thus, we build upon the linear CoT framework by incorporating non-linear structures to simultaneously capture both linear and non-linear human reasoning. Tree-of-thoughts focuses on modeling non-linear thoughts explicitly, whereas our approach leverages non-linear structures to assist the Chain-of-Thought reasoning.

\paragraph{Few-shot CoT Prompting} 

Few-shot CoT reasoning for LLMs, however, utilizes multiple input-output pairs to prompt the LLMs to output CoT and obtain the final answer. Due to its ability to provide better performance compared to Zero-shot CoT, Few-shot CoT has gained more attention in research, particularly through effective demonstrations.
Few-shot CoT prompting was first formally explored by~\citet{DBLP:journals/corr/abs-2201-11903} and is a form of discrete prompt learning that involves context learning in large models. Compared to traditional in-context learning, which prompts LLMs with a list of input-output demonstration pairs along with a test input to allow the model to predict output, Few-shot CoT prompting outputs additional logical reasoning procedures apart from the target output.~\citet{DBLP:journals/corr/abs-2203-11171} proposed a follow-up method to~\cite{DBLP:journals/corr/abs-2201-11903}. The main improvement is that the model uses the majority vote for the answers, which was found to significantly improve the performance of the CoT. However, these few-shot CoT models depend on hand-crafted demonstrations. To solve this problem,~\citet{DBLP:journals/corr/abs-2210-03493} proposed Auto-CoT, which maintains the diversity of sampled questions and generates reasoning chains to automatically construct demonstrations. Specifically, Auto-CoT consists of two main stages: (1) Problem clustering: divide the given dataset of problems into several clusters; (2) Demonstration sampling: select a representative problem from each cluster and use a simple heuristic method to generate its reasoning chain.
Furthermore,~\citet{DBLP:journals/corr/abs-2304-09842} also explores few-shot CoT reasoning for recently popular LLMs ChatGPT and GPT-4.


\paragraph{CoT Fine-tuning} 
In~\citet{DBLP:journals/corr/abs-2302-00923}, it was proposed to fine-tune smaller language models instead of prompting them in LLMs. And this approach enabled the CoT to go beyond textual information and incorporate visual (image) modalities using a gated fusion mechanism into a two-stage CoT. The results demonstrated that CoT fine-tuning with fewer parameters has potential. Therefore, in this work, we focus on fine-tuning for CoT to reduce the number of required model parameters and help LLMs better comprehend different modalities. However, previous CoT research has been limited to different modalities, such as textual and vision information, without considering the deduction reasoning process.
Therefore, in this work, we move beyond modeling the reasoning process solely as a thought chain and elevate it to a thought graph. We provide a more comprehensive and nuanced representation, enabling LLMs to perceive the deduction reasoning process accurately, resulting in more precise answer generation.

\section{Dataset }
\label{ap:data}
 AQUA-RAT dataset consists of about 100,000 algebraic word problems with natural language rationales. For AQUA-RAT, the model is trained to reasoning through the steps to generate the final answer. ScienceQA benchmark is the pioneering large-scale dataset for multimodal science questions, equipped with comprehensive annotations for answers, including detailed lectures and explanations. The dataset contains 21k questions covering three subjects: natural science, language science, and social science. Each question is presented with a context in the form of natural language or an optional image. The model is trained to elucidate the reasoning process in natural language while choosing the answer from a set of options.

\begin{table}[h]
    \centering
        \begin{tabular}{cc}
    \hline
    Splits & \#Problems \\ \hline
    Train  & 97467      \\
    Dev  & 254      \\
    Test   & 254      \\ \hline
    \end{tabular}
    \caption{AQUA-RAT dataset statistics (\# denotes numbers)}
    \label{tab:datasetgsm8k}    \centering
\end{table}

\begin{table}[h]
\centering
\begin{tabular}{ll}
        \hline
        \multicolumn{1}{c}{Statistic} & \multicolumn{1}{c}{Number} \\ \hline
        Splits                        &                            \\ \hline
        \#Train                       & 12,726                      \\
        \#Dev                         & 4,241                       \\
        \#Test                        & 4,241                       \\
        \#Total                       & 21,208                     \\ \hline
        Attribute                     &                            \\ \hline
        \#Subjects                    & 3                          \\
        \#Topic                       & 26                         \\
        \#Category                    & 127                        \\
        \#Skill                       & 379                        \\ \hline
        \end{tabular}
        \caption{ScienceQA dataset statistics (\# denotes numbers)}
        \label{tab:dataset}
\end{table}

\section{Training Parameters}
\label{ap:param}
\begin{table}[h]
    \centering
    \begin{tabular}{lc}
    \hline
    \multicolumn{1}{c}{Parameters} & Value  \\ \hline
    Epochs                         & 100    \\
    Batch size for T5-base (per device)                     & 10     \\
    Batch size for T5-large (per device)                     & 8     \\
    Learning rate      & 5e-5   \\
    Weight decay       & 0.01 \\
    Max input length      & 512   \\
    Max number of nodes                   & 150    \\ \hline
    \end{tabular}
    \caption{Training parameters for GoT}
    \label{param}
\end{table}

\section{Rationale Generation Results}
\label{append:rationale}
The rationale genration results can be found in Table \ref{tab:rationale results}. We can observe from Table \ref{tab:rationale results} that the impact of GoT on rationale generation is limited. We attribute this limitation to the fact that the input text for thought graph construction only includes questions and choices. Consequently, the thought graph constructed from such limited information can only facilitate constrained deductive reasoning. However, in the answer generation stage, when provided with rationales, the model needs to possess stronger deductive reasoning capabilities to understand the relationship between rationales, questions, and choices.
\begin{table*}[h]
\centering
\scalebox{0.8}{
\begin{tabular}{lcccc}
\toprule
MODELS                   & BLEU1 & BLEU4 & ROUGE & SIMILARITY \\ \midrule

\multicolumn{5}{l}{\textit{AQUA-RAT}}                                                     \\
FLAN-Alpaca$_\textrm{base}$         & 19.78                         & 3.49  & 28.40  & 68.61     \\
FLAN-Alpaca$_\textrm{large}$        & 22.45                             & 5.40      & 29.55      & 70.34         \\
\textbf{GoT-T5$_\textrm{base}$}                & 22.05                         & 5.02  & 29.18  & 69.09     \\
\textbf{GoT-T5$_\textrm{large}$}             & \textbf{24.47}                         & \textbf{6.68}  & \textbf{29.86}  & \textbf{71.58}     \\ \midrule
\multicolumn{5}{l}{\textit{ScienceQA}}                                                 \\
Mutimodal-CoT$_\textrm{base}^*$~\cite{DBLP:journals/corr/abs-2302-00923}  & 91.04                         & 86.81  & 93.24  & 96.34     \\ 
\textbf{GoT-T5$_\textrm{base}$}                & 92.50  & 88.79  & 94.39  & 96.74     \\
\textbf{GoT-T5$_\textrm{large}$}              & \textbf{93.49}                         & \textbf{90.09}  & \textbf{95.17}  & \textbf{97.33}     \\ \bottomrule
\end{tabular}
}
\caption{Rationale generation results (\%). (*: we re-run the Mutimodal-CoT$_\textrm{base}$ to report the full rationale scores. We use sentence-transformers (\hyperlink{{https://huggingface.co/sentence-transformers/all-MiniLM-L6-v2}}{{https://huggingface.co/sentence-transformers/all-MiniLM-L6-v2}}) to obtain sentence embeddings and calculate the cosine similarity for SIMILARITY)}
\label{tab:rationale results}
\end{table*}

\section{Case Study}
\label{ap:case}

\begin{figure*}[h]
    \centering
    \includegraphics[width=1\textwidth]{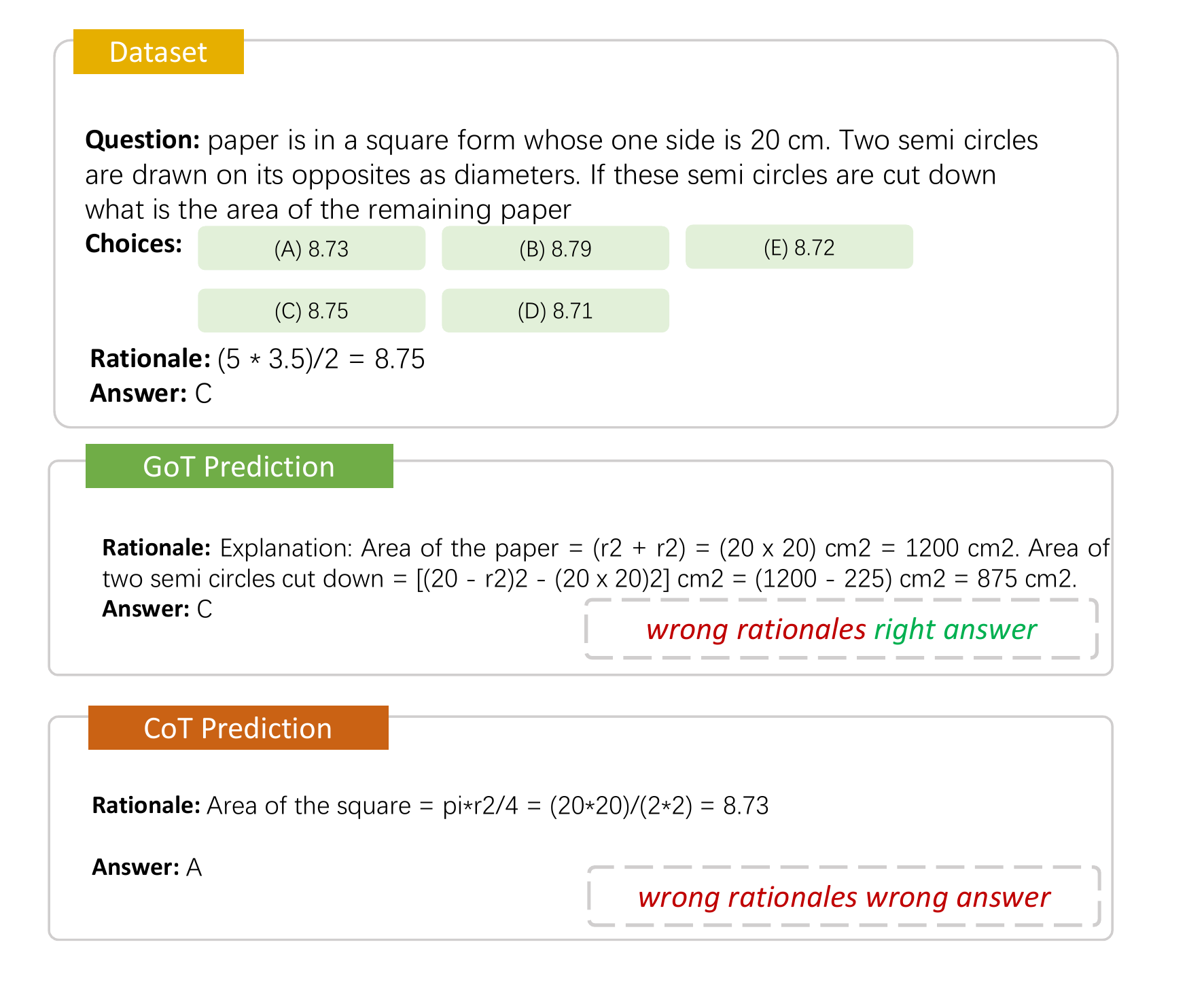}
    \caption{Examples of AQUA-RAT }
    \label{fig:case_aqua}
\end{figure*}

\begin{figure*}[h]
    \centering
    \includegraphics[width=1\textwidth]{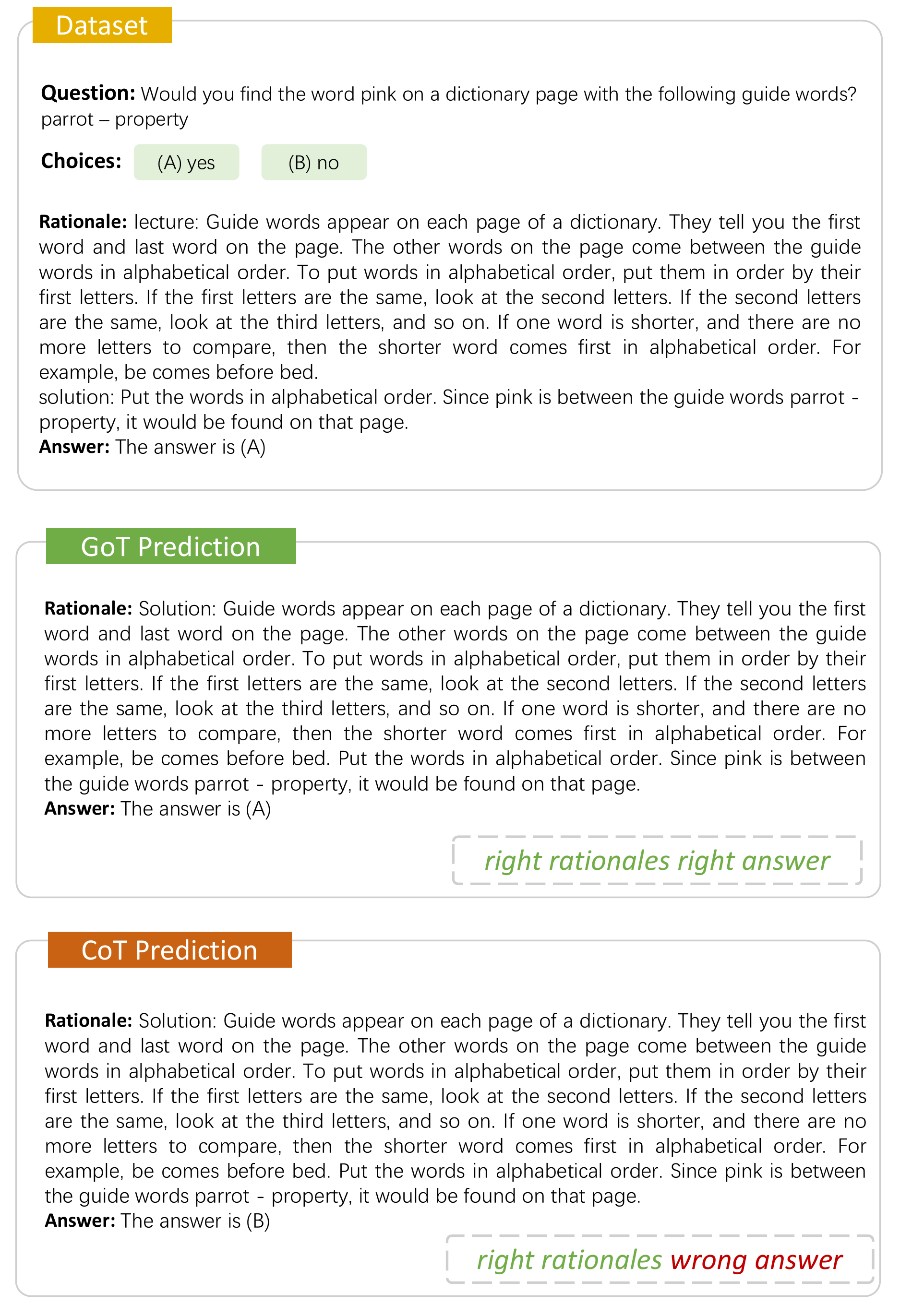}
    \caption{Examples of ScienceQA }
    \label{fig:case4}
\end{figure*}

\begin{figure*}[h]
    \centering
    \includegraphics[width=1\textwidth]{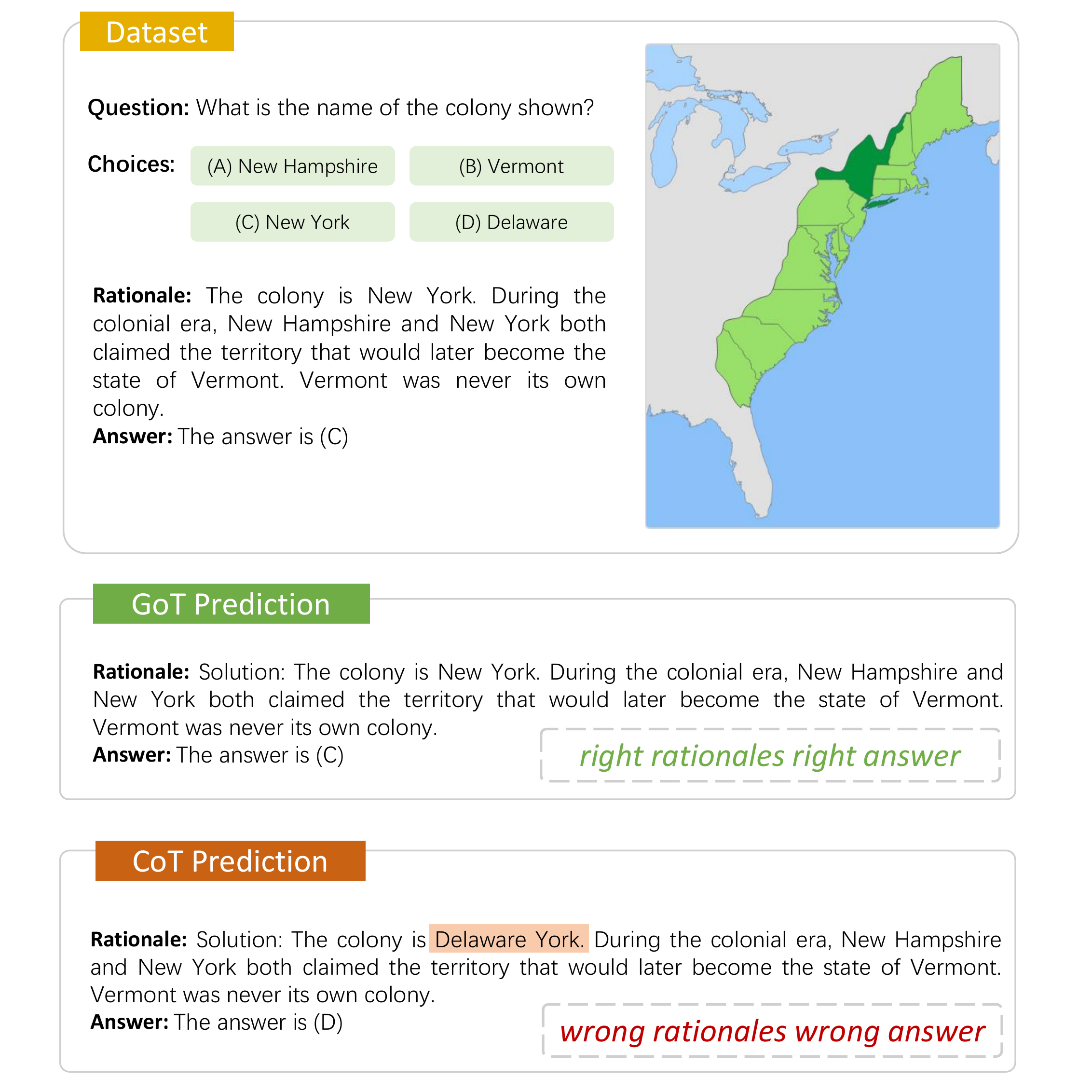}
    \caption{Examples of ScienceQA }
    \label{fig:case1}
\end{figure*}

\begin{figure*}[h]
    \centering
    \includegraphics[width=0.8\textwidth]{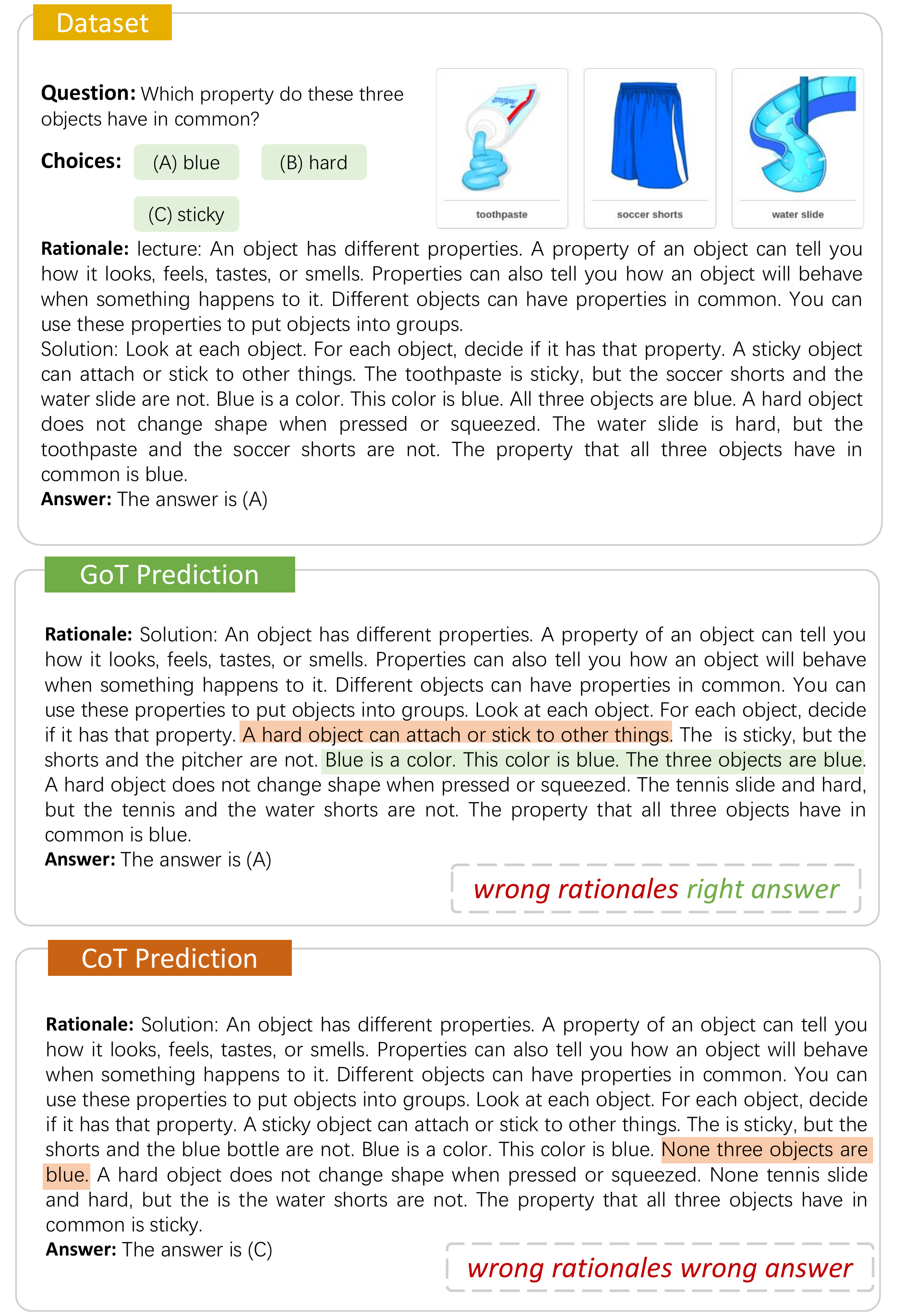}
    \caption{Examples of ScienceQA }
    \label{fig:case3}
\end{figure*}

\begin{figure*}[h]
    \centering
    \includegraphics[width=1\textwidth]{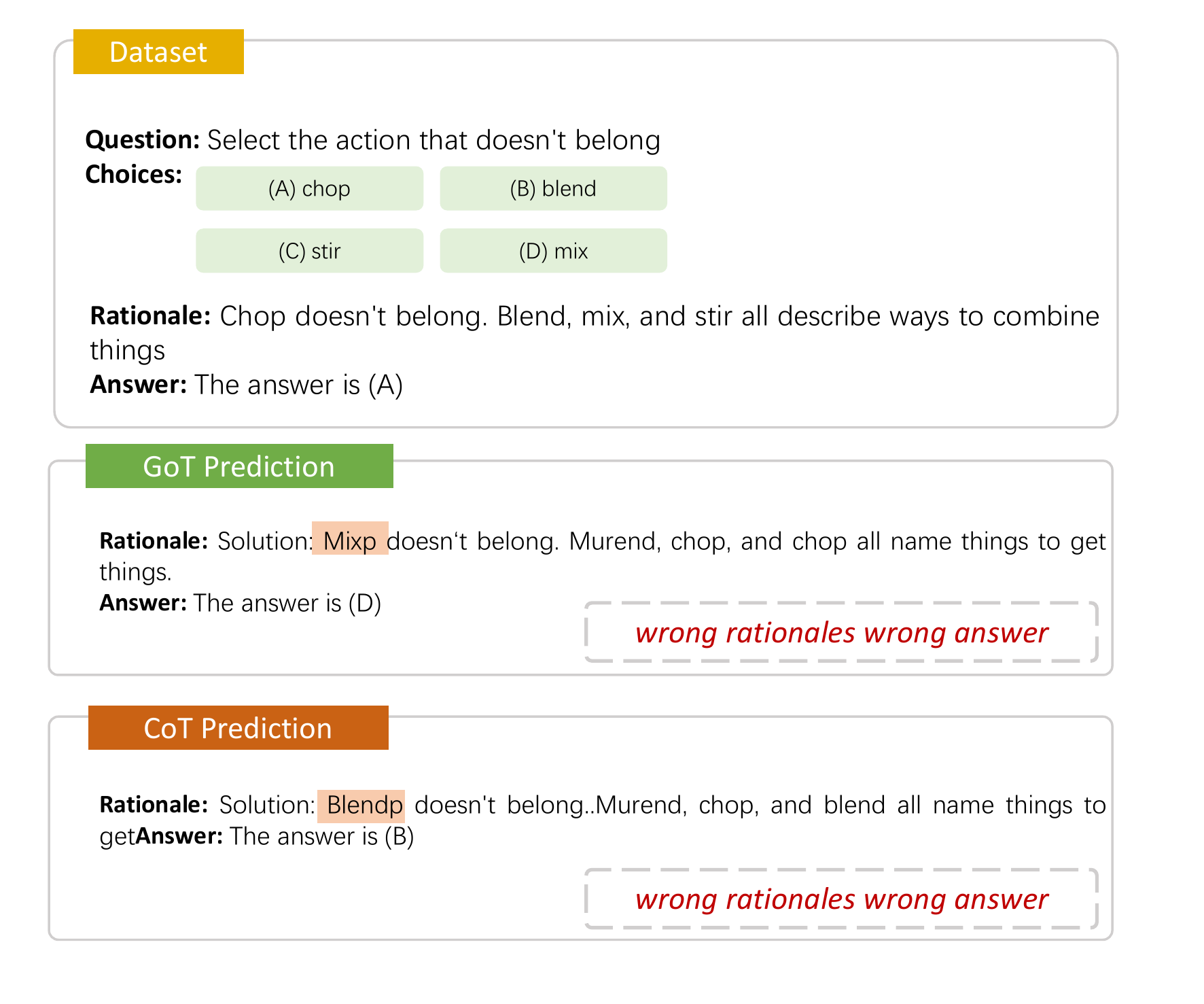}
    \caption{Examples of ScienceQA }
    \label{fig:case2}
\end{figure*}

To facilitate a more illustrative comparison between GoT and the CoT, we have selected several representative examples. Figure \ref{fig:case_aqua} illustrates the examples from AQUA-RAT dataset. Figure \ref{fig:case4} to Figure \ref{fig:case2} illustrates examples from ScienceQA dataset. From Figure \ref{fig:case4} and Figure \ref{fig:case1}, we can see that GoT can better understand the rationales and generate more accurate result. In Figure \ref{fig:case3}, we can see that when provided with wrong rationale, our model is more robust to the noise and can focus on more important key information. (We highlight the noisy wrong rationale in red and correct key rationale in green). Figure \ref{fig:case2} presents a language problem which have less context and requires a certain amount of common sense knowledge. Hence, the impact of constructing a mind map on enhancing the model is not significant. Therefore, both GoT and CoT predict wrong answers.

\section{Representation Visualization}
\begin{figure*}[h]
    \centering
    \includegraphics[width=1\textwidth]{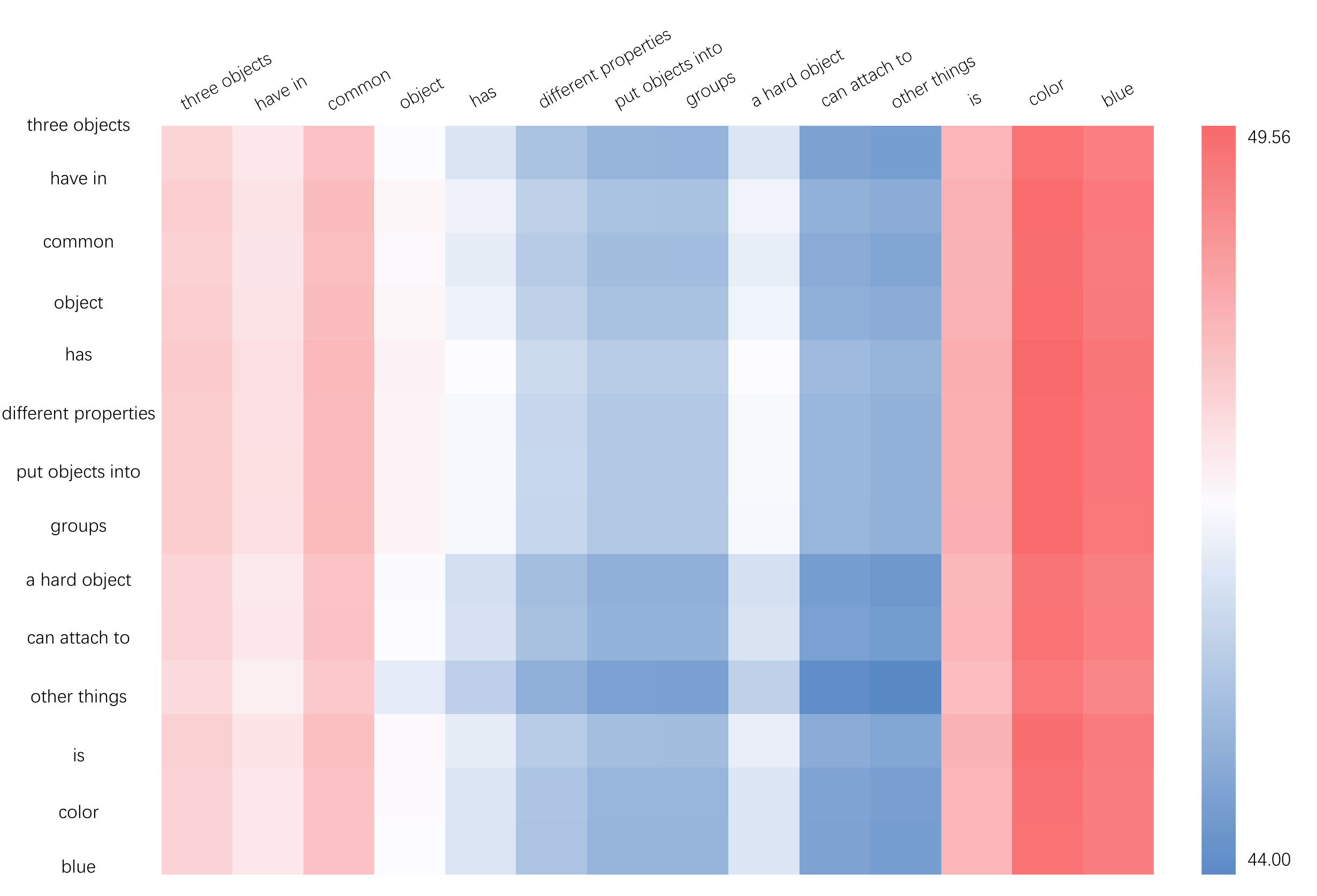}
    \caption{Representation visualization }
    \label{fig:viz}
\end{figure*}

In order to demonstrate the deductive reasoning process of GoT more intuitively, we visualized the attention weights of the GoT encoder. The visualization results can be found in Figure \ref{fig:viz}. We took Figure \ref{fig:case3} as an example. In Figure \ref{fig:case3}, even given a wrong rationale, GoT still manages to generate the right answer. We select 14 representative thought nodes and found that "blue","color", and "common" have the greatest weights which indicates that GoT guides the model to focus on more important words and conduct correct deductive reasoning. For the disruptive node "a hard object," our model can effectively discriminate against it and assign a lower attention weight to prevent the model from selecting incorrect answers, as traditional CoT models often do due to erroneous rationales.

\label{ap:viz}

\section{Limitation}
\label{ap:limiation}
Compared to Mutimodal-CoT~\cite{DBLP:journals/corr/abs-2302-00923}, incorporating GoT may result in additional computational costs and slightly slower training times. The training parameters and inference times of the different models are presented in Table \ref{tab:limitation}, which reveals that our model requires a 0.2\% increase in parameters compared to Mutimodal-CoT.

\begin{table}[h]
\centering
\scalebox{0.8}{
\begin{tabular}{lcc}
\toprule
                  & \#Parameters & \begin{tabular}[c]{@{}c@{}}Inference time \\ (eval samples/per second)\end{tabular} \\ \midrule
Mutimodal-CoT$_\textrm{base}$ & 227M         & 16.33                                                                               \\
Ours              & 233M         & 13.38                                                                               \\ \bottomrule
\end{tabular}
}
\caption{The number of training parameters and inference time of different models (\# denotes numbers)}
\label{tab:limitation}
\end{table}

\end{document}